\documentclass[runningheads]{llncs}
\usepackage{graphicx}
\usepackage{amsmath,amssymb} 
\usepackage{color}
\usepackage{bbding}
\usepackage{multirow}
\usepackage[width=122mm,left=12mm,paperwidth=146mm,height=193mm,top=12mm,paperheight=217mm]{geometry}
\usepackage{booktabs} 
\usepackage{algorithm}
\usepackage{algorithmic}
\usepackage{subfigure}
\usepackage{url}
\usepackage{dsfont}
\usepackage{multirow}
\usepackage[usenames,dvipsnames,table]{xcolor}
\usepackage{graphicx}
\usepackage{amsmath,amssymb} 
\usepackage{color}
\usepackage{bbding}
\usepackage{multirow}
\usepackage{epsfig}

\usepackage{times}
\usepackage{epsfig}
\usepackage{graphicx}
\usepackage{amsmath}
\usepackage{amssymb}
\usepackage{booktabs} 
\usepackage{algorithm}
\usepackage{algorithmic}
\usepackage{subfigure}
\usepackage{url}
\usepackage{multirow}
\usepackage[usenames,dvipsnames,table]{xcolor}

\usepackage{dsfont}
\usepackage{amsmath,amssymb} 
\usepackage{color}

\begin{document}
\pagestyle{headings}
\mainmatter

\title{Peak-Piloted Deep Network for Facial Expression Recognition} 
\authorrunning{X. Zhao, X. Liang, L. Liu, T. Li, Y. Han, N. Vasconcelos, S. Yan}

\author{Xiangyun Zhao$^{1}$ \quad Xiaodan Liang$^{2}$ \quad Luoqi Liu$^{3,4}$ \quad Teng Li$^{5}$\\ \quad Yugang Han$^{3}$ \quad Nuno Vasconcelos$^{1}$ \quad Shuicheng~Yan $^{3,4}$}

\institute{$^{1}$ University of California, San Diego\quad $^{2}$ Carnegie Mellon University \\ \quad $^{3}$ 360 AI Institute \quad $^{4}$ National University of Singapore\\ \quad  $^{5}$ Institute of Automation, Chinese Academy of Sciences\\ \email{ xiz019@ucsd.edu xdliang328@gmail.com liuluoqi@360.cn \\tenglwy@gmail.com hanyugang@360.cn\\ nvasconcelos@ucsd.edu eleyans@nus.edu.sg}}

\maketitle

\begin{abstract}
	
Objective functions for training of deep networks for face-related 
recognition tasks, such as facial expression recognition (FER), usually 
consider each sample independently. In this work, we present a novel 
peak-piloted deep network (PPDN) that uses a sample with peak 
expression (easy sample) to supervise the intermediate feature responses
for a sample of non-peak expression (hard sample) of the same type and from 
the same subject. The expression
evolving process from non-peak expression to peak expression can thus be
implicitly embedded in the network to achieve the invariance to expression intensities. A special-purpose back-propagation procedure, peak gradient
suppression (PGS), is proposed for network training. It drives the 
intermediate-layer feature responses of non-peak expression samples
towards those of the corresponding peak expression samples, while
avoiding the inverse. This avoids degrading the recognition capability for 
samples of peak expression due to interference from their non-peak expression 
counterparts. Extensive comparisons on two popular FER datasets, Oulu-CASIA
and CK+, demonstrate the superiority of the PPDN over state-of-the-art 
FER methods, as well as the advantages of both the network structure and 
the optimization strategy. Moreover, it is shown that PPDN is
a general architecture, extensible to other tasks by proper definition of 
peak and non-peak samples. This is validated by experiments that
show state-of-the-art performance on pose-invariant face recognition,
using the Multi-PIE dataset.


\end{abstract}

\section{Introduction}

\begin{figure}
	\centering
	\includegraphics[scale=0.42]{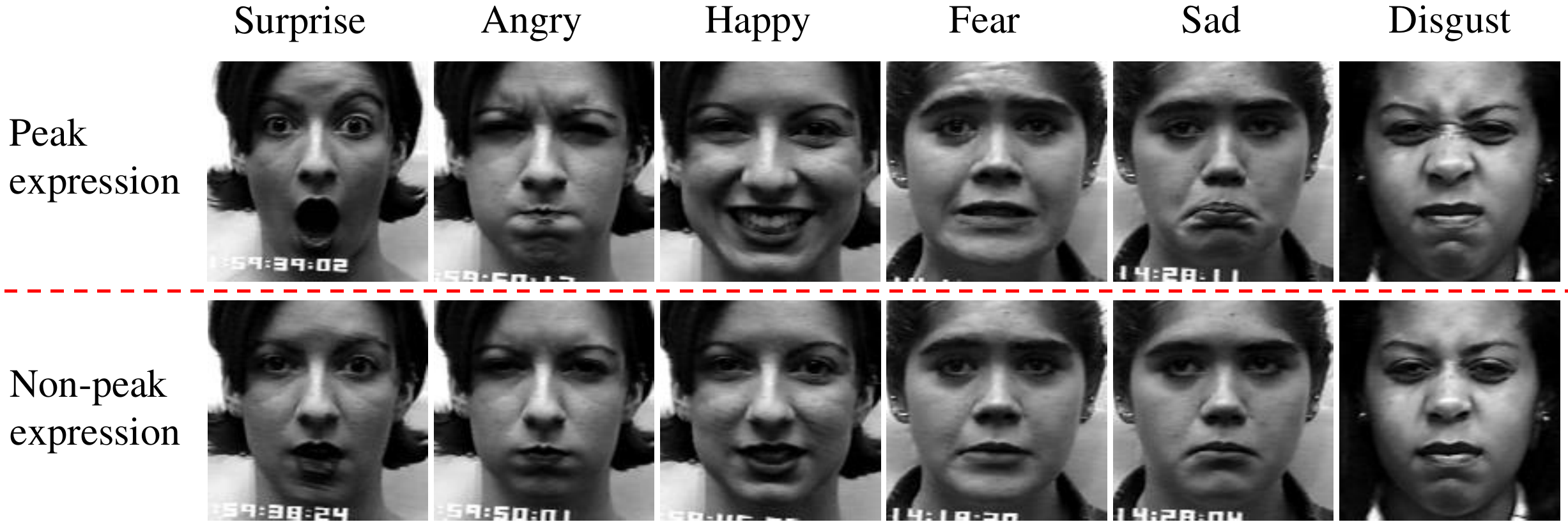}
	\caption{Examples of six facial expression samples, including surprise, angry, happy, fear, sad and disgust. For each subject, the peak and non-peak expressions are shown. }
	\label{fig:expression}
\end{figure}

Facial Expression Recognition~(FER) aims to predict the basic facial 
expressions (e.g. happy, sad, surprise, angry, fear, disgust) from a human 
face image, as illustrated in Fig.~\ref{fig:expression}.\footnote{This work was performed when Xiaoyun Zhao was an intern at 360 AI Institute.}
 Recently, FER has 
attracted much research 
attention~\cite{liu2014learning,chen20153d,dapogny2015pairwise,liu2014facial,yu2015image,liu2013aware,jung2015joint}. It can facilitate other face-related tasks, such as face 
recognition~\cite{li2006expression} and alignment~\cite{zhang2014facial}. 
Despite significant recent progress
\cite{zhong2012learning,shan2009facial,liu2014facial,kahou2014facial}, FER 
is still a challenging problem, due to the following 
difficulties. First, as illustrated 
in Fig.~\ref{fig:expression}, different subjects often dsiplay the same 
expression with diverse intensities and visual appearances. In
a videostream, an expression will first appear in a subtle form 
and then grow into a strong display of the underlying feelings. 
We refer to the former as a non-peak and to the latter as a peak expression. 
Second, peak and non-peak expressions by the same subject can have 
significant variation in terms of attributes such as mouth corner radian, 
facial wrinkles, etc. 
Third, non-peak expressions are more commonly displayed than peak 
expressions. It is usually difficult to capture critical and subtle
expression details from non-peak expression images, which can be hard to 
distinguish across expressions. For example, the non-peak expressions for fear 
and sadness are quite similar in Fig.~\ref{fig:expression}.

Recently, deep neural network
architectures have shown excellent performance in face-related recognition 
tasks~\cite{chopra2005learning,lai2015deep,li2015convolutional}. The has
led to the introduction of FER network 
architectures~\cite{liu2014facial,mollahosseini2015going}. There are, nevertheless, some important limitations. First,
most methods consider each sample independently during learning,
ignoring the intrinsic correlations between each pair of samples (e.g., easy 
and hard samples). This limits the discriminative capabilities of the learned 
models. Second, they focus on recognizing the clearly separable peak 
expressions and ignore the most common non-peak expression samples, whose
discrimination can be extremely challenging.

\begin{figure}[t]
\centering
\includegraphics[scale=0.42]{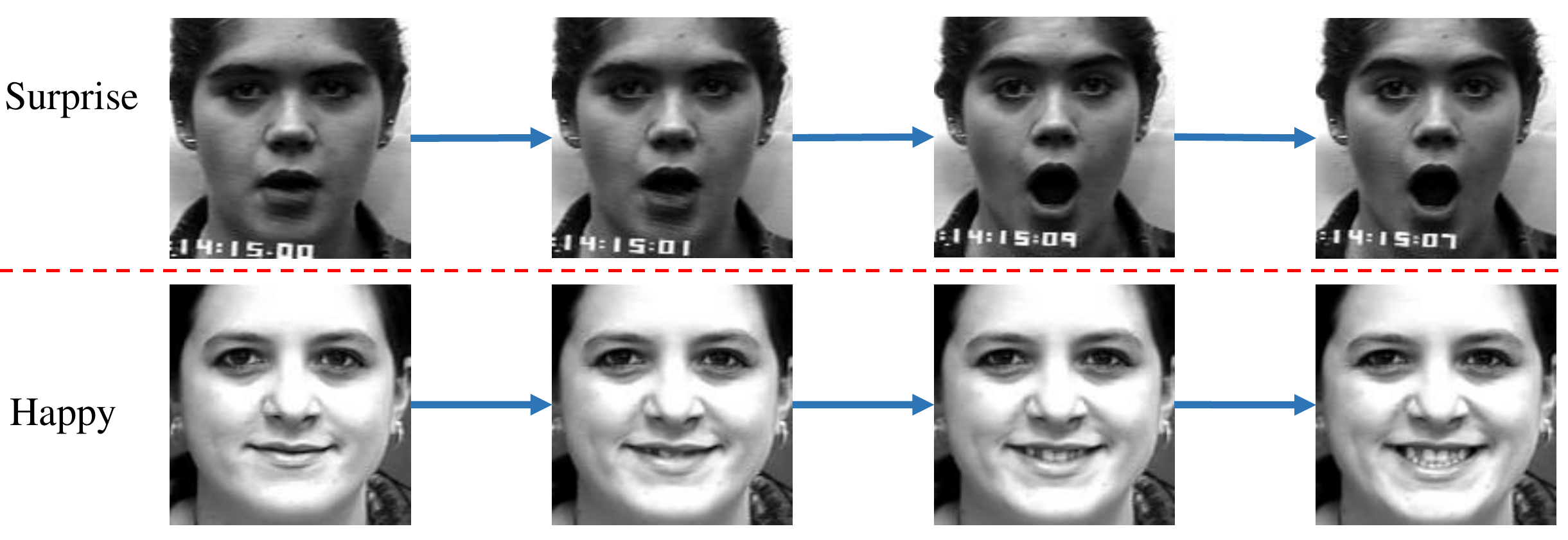}
\caption{Expression evolving process from non-peak expression to peak expression. 
	}
\label{fig:evolve}
\end{figure}

In this paper, we propose a novel peak-piloted deep network (PPDN) 
architecture, which  implicitly embeds the natural evolution of expressions
from non-peak to peak expression in the learning 
process, so as to zoom in on the subtle differences between weak expressions
and achieve invariance to expression intensity.  Intuitively, 
as illustrated in Fig.~\ref{fig:evolve}, peak and non-peak expressions from 
the same subject often exhibit very strong visual correlations (e.g., similar 
face parts) and can mutually help the recognition of each other. The 
proposed PPDN uses the feature responses to samples of peak expression 
(easy samples) to supervise the responses to samples of non-peak 
expression (hard samples) of the same type and from the same subject. 
The resulting mapping of non-peak expressions into their corresponding 
peak expressions magnifies their critical and subtle details, facilitating
their recognition.

In principle, an explicit mapping from non-peak to peak expression 
could significantly improve recognition. However, such a mapping is
challenging to generate, since the detailed changes of face features 
(e.g., mouth corner radian and wrinkles) can be quite difficult to predict. 
We avoid this problem by focusing on 
the high-level feature representation of the facial expressions, which
is both more abstract and directly related to facial expression recognition. 
In particular, the proposed PPDN optimizes the tasks of 1) feature 
transformation from non-peak to peak expression and 2) recognition of facial 
expressions in a unified manner. It is, in fact, a general approach,
applicable to many other recognition tasks (e.g. face recognition) by 
proper definition of peak and non-peak samples (e.g. frontal and 
profile faces). By implicitly learning the evolution from hard poses (e.g., profile faces) to  easy poses 
(e.g., near-frontal faces), it
can improve the recognition accuracy of prior solutions to these
problems, making them more robust to pose variation.

During training, the PPDN takes an image pair with a peak and a non-peak 
expression of the same type and from the same subject. 
This image pair is passed through several intermediate layers to generate 
feature maps for each expression image. The L2-norm of the difference 
between the feature maps of non-peak and peak expression images is then 
minimized, to embed the evolution of expressions into the PPDN framework. 
In this way, the PPDN incorporates the peak-piloted feature transformation 
and facial expression recognition into a unified architecture. The PPDN
is learned with a new back-propagation algorithm, denotes peak gradient 
suppression (PGS), which drives the feature responses to non-peak 
expression instances towards those of the corresponding peak expression 
images, but not the contrary. This is unlike the traditional optimization of
Siamese networks~\cite{chopra2005learning}, which encourages the feature 
pairs to be close to each other, treating the feature maps of the two images 
equally. Instead, the PPDN focuses on transforming the features 
of non-peak expressions towards those of peak expressions. 
This is implemented by, during each back-propagation iteration, 
ignoring the gradient information due to the peak expression image in 
the L2-norm minimization of feature differences, while keeping that
due to the non-peak expression. The gradients of the recognition loss,
for both peak and non-peak expression images, are the same as in traditional 
back-propagation. This avoids the degradation of the recognition 
capability of the network for samples of peak expression due to the
influence of non-peak expression samples.

Overall, this work has four main contributions. 1)~The PPDN architecture 
is proposed, using the responses to samples of peak expression (easy samples) 
to supervise the responses to samples of non-peak expression (hard 
samples) of the same type and from the same subject. The targets of 
peak-piloted feature transformation and facial expression recognition, for 
peak and non-peak expressions, are optimized  simultaneously. 
2) A tailored back-propagation procedure, PGS, is proposed to drive the 
responses to non-peak expressions towards those of the corresponding peak 
expressions, while avoiding the inverse. 3) The PPDN is shown to perform
intensity-invariant facial expression recognition, by effectively 
recognizing the most common non-peak expressions. 4) Comprehensive evaluations 
on several FER datasets, 
namely CK+~\cite{lucey2010extended} and Oulu-CASIA~\cite{zhao2011facial}, 
demonstrate the superiority of the PPDN over previous methods.
Its generalization to other tasks is also demonstrated through
state-of-the-art robust face recognition performance on the public 
Multi-PIE dataset~\cite{gross2010multi}.

\section{Related Work}

There have been several recent attempts to solve the facial expression 
recognition problem. These methods can be grouped into two categories: 
sequence-based and still image approaches. In the first category, 
sequence-based approaches~\cite{jung2015joint,liu2014learning,guo2012dynamic,zhao2011facial,klaser2008spatio}  exploit both the appearance and motion 
information from video sequences. In the second category, still image 
approaches~\cite{zhong2012learning,liu2014facial,kahou2014facial} recognize 
expressions uniquely from image appearance patterns. Since still image 
methods are more generic, recognizing 
expressions in both still images and sequences, we focus on models for 
still image expression recognition. Among these, both hand-crafted pipelines 
and deep learning methods have been explored for FER. Hand-crafted 
approaches~\cite{zhong2012learning,sikka2012exploring,shan2009facial} perform three steps sequentially: feature extraction, feature selection 
and classification. This can lead to suboptimal recognition, due to 
the combination of different optimization targets. 

Convolutional Neural Network (CNN) 
architectures~\cite{krizhevsky2012imagenet,szegedy2015going,simonyan2014very} 
have recently shown excellent performance on face-related recognition 
tasks~\cite{sun2014deep,zhu2013deep,yim2015rotating}. Methods that resort to the 
CNN architecture have also been proposed for FER. For example, Yu et 
al.~\cite{yu2015image} used an ensemble of multiple deep
CNNs. Mollahosseini et 
al.~\cite{mollahosseini2015going} used three inception 
structures~\cite{szegedy2015going} in convolution for FER. All these methods
treat expression instances of different intensities of the same subject 
independently. Hence, the correlations between peak and non-peak 
expressions are overlooked during learning. In contrast, the proposed PPDN 
learns to embed the evolution from non-peak to peak expressions, so as 
to facilitate image-based FER.

\begin{figure}[t]
	\centering
	\includegraphics[scale=0.36]{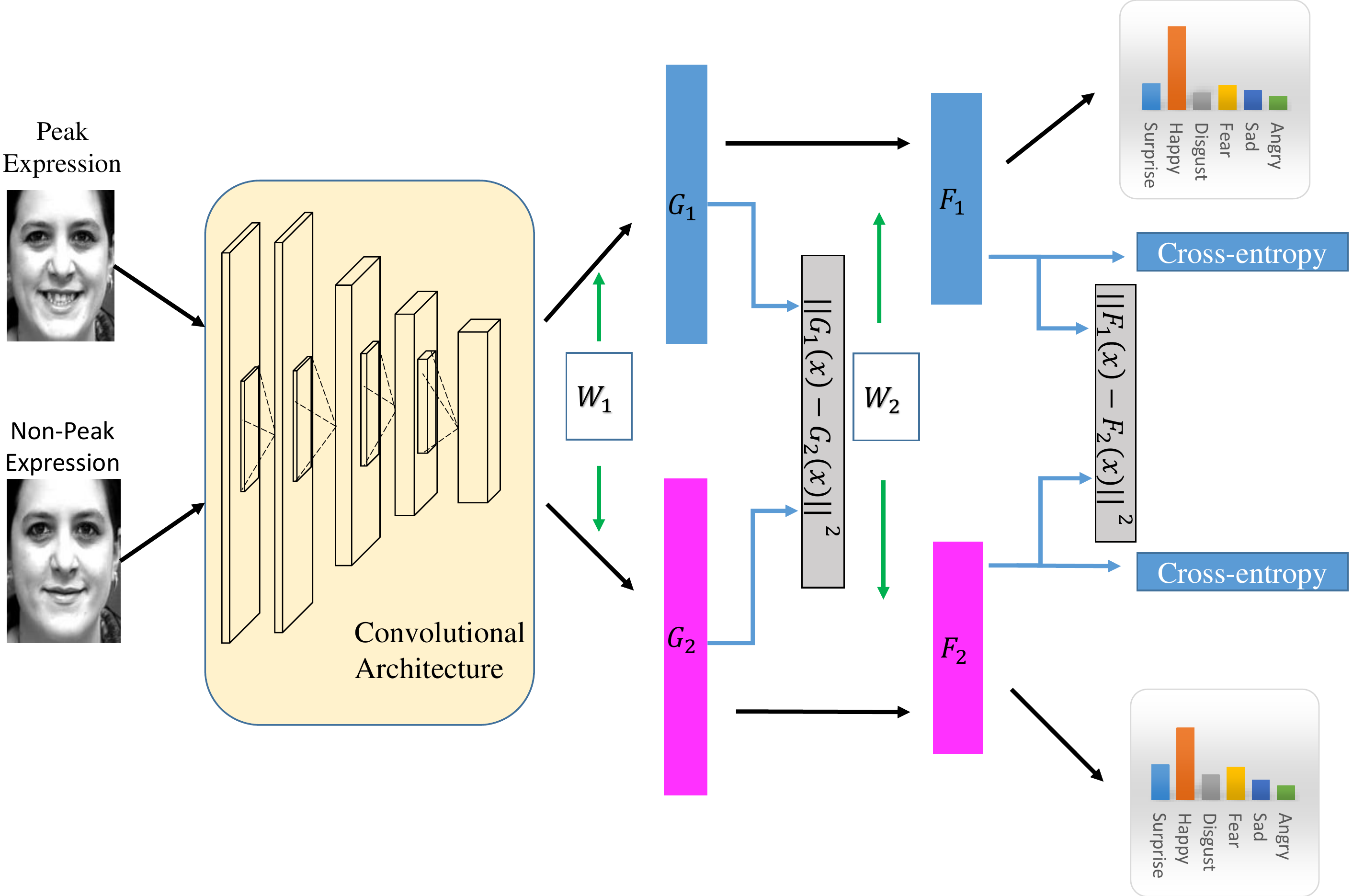}
	\caption{Illustration of the training stage of PPDN. During training, PPDN takes the pair of peak and non-peak expression images as input. After passing the pair through several convolutional and fully-connected layers, the intermediate feature maps can be obtained for peak and non-peak expression images, respectively. The L2-norm loss between these feature maps is optimized for driving the features of the non-peak expression image towards those of the peak expression image. The network parameters can thus be updated by jointly optimizing the L2-norm losses and the losses of recognizing two expression images. During the back-propagation process, the Peak Gradient Suppression (PGS) is utilized.}
	\label{fig:architecture}
\end{figure}

\section{The Peak-Piloted Deep Network (PPDN)}

In this work we introduce the PPDN framework, which implicitly learns 
the evolution from non-peak to peak expressions, in
the FER context. As illustrated in 
Fig.~\ref{fig:architecture}, during training the PPDN takes an image pair as 
input. This consists of a peak and a non-peak expression of the same 
type and from the same subject. This image pair is passed through several 
convolutional and fully-connected layers, generating pairs of feature 
maps for each expression image. To drive the feature responses to the 
non-peak expression image towards those of the peak expression image, the 
L2-norm of the feature differences is minimized. 
The learning algorithm optimizes a combination of this L2-norm loss
and two recognition losses, one per expression image. 
Due to its excellent performance on several face-related recognition 
tasks~\cite{schroff2015facenet,sun2015deepid3}, the popular 
GoogLeNet~\cite{szegedy2015going} is adopted as the basic 
network architecture. The incarnations of the inception architecture in 
GoogLeNet are restricted to filters sizes $1 \times 1$, $3 \times 3$ and 
$5 \times 5$. In total, the GoogLeNet implements nine inception structures 
after two convolutional layers and two max pooling layers. After that, the 
first fully-connected layer produces the intermediate features with 1024 
dimensions, and the second fully-connected layer generates the label 
predictions for six expression labels. During testing, the PPDN takes one 
still image as input, outputting the predicted probabilities for all six 
expression labels. 

\subsection{Network Optimization}

The goal of the PPDN is to learn the evolution from non-peak 
to peak expressions, as well as recognize the basic facial expressions. 
We denote the training set as 
$S =  \lbrace x^{p}_{i},x^n_{i},y^p_{i},y^n_i,i = 1,...,N \rbrace$, where 
sample $x^n_{i}$ denotes a face with non-peak expression, $x^p_{i}$ a
face with the corresponding peak expression, and $y^n_{i}$ and $y^p_{i}$ are 
the corresponding expression labels. To supervise the feature responses
to the non-peak expression instance with those of the peak expression instance, 
the network is learned with a loss function that includes the L2-norm 
of the difference between the feature responses to peak and non-peak 
expression instances. Cross-entropy losses are also used to optimize the 
recognition of the two expression images. Overall, the loss of the PPDN is
\begin{equation}\begin{aligned} J &= \frac{1}{N}(J_1 + J_2 +J_3 + \lambda \sum_{i=1}^N||W||^2) \\
& = \frac{1}{N}\sum_{i=1}^N \sum_{j \in \Omega}\Vert f_j(x^p_{i},W) - f_j(x^n_{i},W) \Vert^2  +  \frac{1}{N}\sum_{i=1}^N L(y^p_i,f(x^p_i;W)) \\
&+ \frac{1}{N}\sum_{i=1}^N L(y^n_i,f(x^n_i;W))+ \lambda||W||^2,
\end{aligned}
\label{eq:loss}
 \end{equation}
where $J_1$, $J_2$ and $J_3$ indicate the L2-norm of the feature differences 
and the two cross-entropy losses for recognition, respectively. Note that the 
peak-piloted feature transformation is quite generic and could be applied to 
the features produced by any layers. We denote $\Omega$ as the set of layers 
that employ the peak-piloted transformation, and $f_j, j \in \Omega$ as the 
feature maps in the j-th layer. To reduce the effects caused by scale 
variability of the training data, the features $f_j$  are L2 normalized
before the L2-norm of the difference is computed. More specifically, the 
feature maps $f_j$ are concatenated into one vector, which is L2 normalized. 
In the second and third terms, $L$ represents the cross-entropy loss 
between the ground-truth labels and the predicted probabilities of all 
labels. The final regularization term is used to penalize the complexity of 
network parameters $W$. Since the evolution from non-peak to 
peak expression is embedded into the network, the latter learns a more 
robust expression recognizer. 
 
\subsection{Peak Gradient Suppression (PGS)}

To train the PPDN, we propose a special-purpose back-propagation algorithm 
 for the optimization of~(\ref{eq:loss}). Rather than the traditional 
straightforward application of stochastic gradient 
descent~\cite{chopra2005learning}~\cite{schroff2015facenet}, the goal is to 
drive the intermediate-layer responses of non-peak expression instances 
towards those of the corresponding peak expression instances, while avoiding 
the reverse. Under traditional stochastic gradient decent 
(SGD)~\cite{bottou2010large}, the network parameters would be updated
with
\begin{equation}
\begin{split}
W^{+} &= W - \gamma \nabla_W J(W; x^p_{i},x^p_{i},y^n_{i},y^p_i)\\
  &= W -\frac{\gamma}{N}\frac{\partial J_{1}(W; x^n_{i},x^p_{i})}{\partial f_{j}(W;x^n_{i})}\times\frac{\partial f_{j}(W;x^n_{i})}{\partial W} - \frac{\gamma
}{N}\frac{\partial J_{1}(W; x^n_{i},x^p_{i})}{\partial f_{j}(W;x^p_{i})}\times\frac{\partial f_{j}(W;x^p_{i})}{\partial W} \\
 & - \frac{1}{N}\gamma \nabla_W J_{2}(W; x^p_{i},y^p_{i})
 - \frac{1}{N}\gamma \nabla_W J_{3}(W;x^n_{i},y^n_{i})
  - 2\gamma W,
\label{eq:SGD}
\end{split}
\end{equation}
where $\gamma$ is the learning rate. The proposed
peak gradient suppression (PGS) learning algorithm
uses instead the updates
 \begin{equation}
 \begin{split}
W^{+} 
  &= W -\frac{\gamma}{N}\frac{\partial J_{1}(W; x^n_{i},x^p_{i})}{\partial f_{j}(W;x^n_{i})}\times\frac{\partial f_{j}(W;x^n_{i})}{\partial W}  \\
 & - \frac{1}{N}\gamma \nabla_W J_{2}(W; x^p_{i},y^p_{i})
 - \frac{1}{N}\gamma \nabla_W J_{3}(W;x^n_{i},y^n_{i})
  - 2\gamma W.
 \label{eq:PGS}
 \end{split}
 \end{equation}
The difference between~(\ref{eq:PGS}) and~(\ref{eq:SGD}) is that the 
gradients due to the feature responses of the peak expression image, 
$-\frac{\gamma}{N}\frac{\partial J_{1}(W; x^n_{i},x^p_{i})}{\partial f_{j}(W;x^p_{i})}\times\frac{\partial f_{j}(W;x^p_{i})}{\partial W}$ are suppressed 
in~(\ref{eq:PGS}). In this way, PGS drives the feature responses of
non-peak expressions towards those of peak expressions, but not the
contrary. In the appendix, we show that this does not prevent learning,
since the weight update direction of PGS  is a descent direction of
the overall loss, although not a steepest descent direction.
\section{Experiments}

To evaluate the PPDN, we conduct extensive experiments on two popular 
FER datasets: CK+~\cite{lucey2010extended} and 
Oulu-CASIA~\cite{zhao2011facial}. To further demonstrate that the PPDN
generalizes to other recognition tasks, we also evaluate its performance
on face recognition over the public Multi-PIE 
dataset~\cite{gross2010multi}.

\subsection{Facial Expression Recognition}
\subsubsection{Training.} The PPDN uses the GoogLeNet~\cite{szegedy2015going} 
as basic network structure. The peak-piloted feature transformation is only 
employed in the last two fully-connected layers. Other configurations,
using the peak-piloted feature transformation on various convolutional 
layers are also reported. Since it is not feasible to train the deep network 
on the small FER datasets available, we pre-trained 
GoogLeNet~\cite{szegedy2015going} on a large-scale face recognition dataset,  
the CASIA Webface dataset~\cite{yi2014learning}. This network was then 
fine-tuned for FER. The CASIA Webface dataset contains 494,414 training 
images from 10,575 subjects, which were used to pre-train the network 
for 60 epochs with an initial learning rate of 0.01.
  For fine-tuning, 
the face region 
was first aligned with the detected eyes and mouth positions.
 The face regions 
were then resized to $128 \times 128$. The PPDN takes a pair of peak and 
non-peak expression images as input. The convolutional layer weights
were initialized with those of the pre-trained model. The weights of the 
fully connected layer were initialized randomly using the 
``xaiver'' procedure~\cite{glorot2010understanding}. The learning rate of 
the fully connected layers was set to 0.0001 and that of pre-trained 
convolutional layers to 0.000001.  ALL models were trained 
using a batch size of 128 image pairs and a weight 
decay of 0.0002. The final trained model was obtained after 20,000 
iterations. For fair comparison with previous 
methods~\cite{zhong2012learning,shan2009facial,liu2014facial}, we did not 
use any data augmentation in our experiments.

\subsubsection{Testing and Evaluation Metric.}

In the testing phase, the PPDN takes one testing image as the input 
and produces its predicted facial expression label. Following the 
standard setting of~\cite{zhong2012learning,shan2009facial}, 10-fold 
subject-independent cross-validation was adopted for evaluation in all 
experiments. 

\subsubsection{Datasets.} FER datasets usually provide video sequences for 
training and testing the facial expression recognizers. We conducted all 
experiments on two popular datasets, CK+~\cite{lucey2010extended} and  
Oulu-CASIA dataset~\cite{zhao2011facial}. For each sequence, the face 
often gradually evolves from a neutral to a peak facial expression. 
CK+ includes six basic facial expressions (angry, happy, surprise, sad, 
disgust, fear) and one non basic expression (contempt).  It contains 
593 sequences from 123 subjects, of which only 327 are annotated with 
expression labels. Oulu-CASIA contains 480 sequences of six facial 
expressions under normal illumination, including 80 subjects between 23 
and 58 years old. 

\begin{table}[!tp]
	\begin{center}
		\caption{Performance comparisons on six facial expressions with four state-of-the-art methods and the baseline using GoogLeNet in terms of average classification accuracy by the 10-fold cross-validation evaluation on CK+ database.}
		\label{table:ckcom_image}
		\begin{tabular}{c|c}
			\hline\noalign{\smallskip}
			Method & Average Accuracy\\
			
			\hline
			
			CSPL~\cite{zhong2012learning}  & 89.9\% \\
			AdaGabor~\cite{bartlett2005recognizing}&  93.3\% \\
			LBPSVM~\cite{shan2009facial} &  95.1\%\\
			BDBN~\cite{liu2014facial}& 96.7\% \\
			GoogLeNet(baseline) & 95.0\% \\
			\hline
			PPDN& \textbf{97.3\%} \\
			\hline
		\end{tabular}
	\end{center}
\end{table}

\begin{table}[!tp]
	\begin{center}
		\caption{Performance comparisons on six facial expressions with UDCS method and the baseline using GoogLeNet in terms of average classification accuracy under same setting as UDCS.}
		\label{table:oulucom_image}
	
		\begin{tabular}{c|c}
		   
			\hline\noalign{\smallskip}
			Method & Average Accuracy\\			
			\hline			
            UDCS~\cite{UDCS} & 49.5\% \\
			GoogLeNet(baseline) & 66.6\% \\
			\hline
			PPDN& \textbf{72.4\%} \\
			\hline
		\end{tabular}
	\end{center}
\end{table}
\subsubsection{Comparisons with Still Image-based Approaches.}  
Table~\ref{table:ckcom_image} compares the PPDN to still image-based 
approaches on CK+, under the standard setting in which only the last
one to three frames (i.e., nearly peak expressions) per sequence are considered 
for training and testing. Four state-of-the-art methods are considered: 
common and specific patches learning (CSPL)~\cite{zhong2012learning},
which employs multi-task learning for feature selection,
AdaGabor~\cite{bartlett2005recognizing} and LBPSVM~\cite{shan2009facial},
which are based on AdaBoost~\cite{AdaBoost}, and Boosted Deep Belief 
Network (BDBN)~\cite{liu2014facial}, which jointly optimizes feature 
extraction and feature selection. In addition, we also compare the PPDN 
to the  baseline ``GoogLeNet (baseline),'' which optimizes the standard 
GoogLeNet with SGD. Similarly to previous 
methods~\cite{zhong2012learning,shan2009facial,liu2014facial},
 the PPDN is evaluated on the last three frames of each sequence. 
Table~\ref{table:oulucom_image} compares the PPDN with UDCS~\cite{UDCS} on 
Oulu-CASIA, under a similar setting where the first 9 images of each 
sequence are ignored, the first 40 individuals are taken as training samples 
and the rest as testing. In all cases, the PPDN input is the pair of one of 
the non-peak frames (all frames other than the last one) and the 
corresponding peak frame (the last frame) in a sequence. 
The PPDN significantly outperforms all other, 
achieving 97.3\% vs a previous best of 96.7\% on 
CK+ and 72.4\% vs 66.6\% on Oulu-CASIA.  This demonstrates the 
superiority of embedding the expression evolution in the network learning.

\subsubsection{Training and Testing with More Non-peak Expressions.} 
The main advantage of the PPDN is its improved ability to
recognize non-peak expressions. To test this,
we compared how performance varies with the number of non-peak expressions. 
Note that for each video sequence, the face expression evolves from neutral 
to a peak expression. The first six frames within a sequence are usually
neutral, with the peak expression appearing in the final frames. 
Empirically, we determined that the 7th to 9th frame often show 
non-peak expressions with very weak intensities, which we
denote as ``weak expressions.'' In addition to the training images used
in the standard setting,  we used all frames beyond the 7th for training. 

Since the previous methods did not publish their codes, we only compare 
the PPDN to the baseline ``GoogLeNet (baseline)''.
Table~\ref{table:ck_compare} reports results for CK+ and 
Table~\ref{table:oulu_compare} for Oulu-CASIA. Three different test sets
were considered: ``weak expression'' indicates that the test set only 
contains the non-peak expression images from the 7th to the 9th frames;
``peak expression'' only includes the last frame; and ``combined'' 
uses all frames from the 7th to the last. ``PPDN (standard SGD)'' is the 
version of PPDN trained with standard SGD optimization, and 
``GoogLeNet (baseline)'' the basic GoogLeNet, taking each expression image 
as input and trained with SGD. The most substantial improvements 
are obtained on the ``weak expression'' test set, $83.36\%$ and
$67.95\%$ of ``PPDN'' vs. $78.10\%$ and $64.64\%$ of ``GoogLeNet (baseline)''
on CK+ and Oulu-CASIA, respectively. This is evidence in support
of the advantage of explicitly learning the evolution from non-peak 
to peak expressions. In addition, the PPDN outperforms ``PPDN (standard SGD)''
and ``GoogLeNet (baseline)'' on the combined sets, where both peak and 
non-peak expressions are evaluated. 

\begin{table}[!tp]\setlength{\tabcolsep}{0.5pt}
	\begin{center}
		\caption{Performance comparison on CK+ database in terms of average classification accuracy of the 10-fold cross-validation when evaluating on three different test sets, including ``weak expression", ``peak expression" and ``combined", respectively.}
		\label{table:ck_compare}
		\begin{tabular}{c|c|c|c}
			\hline\noalign{\smallskip}
			Method & weak expression & peak expression & combined\\
			\hline
			PPDN(standard SGD) & 81.34\% & 99.12\% & 94.18\%\\
			GoogLeNet (baseline)   & 78.10\% & 98.96\% & 92.19\%\\
			\hline
			PPDN & \textbf{83.36\%} & \textbf{99.30\%} & \textbf{95.33\%}\\
			\hline
		\end{tabular}
	\end{center}
\end{table}

\begin{table}[!tp]\setlength{\tabcolsep}{0.5pt}
	\begin{center}
		\caption{Performance comparison on Oulu-CASIA database in terms of average classification accuracy of the 10-fold cross-validation when evaluating on three different test sets, including ``weak expression", ``peak expression" and ``combined", respectively.}
		\label{table:oulu_compare}
		\begin{tabular}{c|c|c|c}
			\hline\noalign{\smallskip}
			Method & weak expression & peak expression & combined\\
			\hline
			PPDN(standard SGD) &  67.05\% & 82.91\% &73.54\%\\	
			GoogLeNet (baseline) & 64.64\%& 79.21\% &71.32\%\\
			\hline
			PPDN  & \textbf{67.95\%}&\textbf{84.59\%} & \textbf{74.99\%}\\
			\hline
		\end{tabular}
	\end{center}
\end{table}

\begin{table}[t]
	\begin{center}
		\caption{Performance comparisons with three sequence-based approaches and the baseline ``GoogLeNet (baseline)" in terms of average classification accuracy of the 10-fold cross-validation on CK+ database. }
		\label{table:ck_sequence}
		\begin{tabular}{c|c|c}
			\hline\noalign{\smallskip}
			Method & Experimental Settings & Average Accuracy\\
			\hline
			3DCNN-DAP~\cite{liu2014deeply}&sequence-based&92.4\%\\
			STM-ExpLet~\cite{liu2014learning}& sequence-based & 94.2\%\\
			DTAGN(Joint)~\cite{jung2015joint} & sequence-based & 97.3\%\\
			GoogLeNet (baseline) & image-based & 99.0\%\\
			PPDN (standard SGD) & image-based & 99.1\%\\
			\hline
			PPDN w/o peak & image-based & \textbf{99.2\%}\\
			PPDN  & image-based & \textbf{99.3\%}\\
			\hline
		\end{tabular}
	\end{center}
\end{table}

\begin{table}[!tp]
	\begin{center}
		\caption{Performance comparisons with five sequence-based approaches and the baseline ``GoogLeNet (baseline)" in terms of average classification accuracy of the 10-fold cross-validation  on Oulu-CASIA.}
		\label{table:oulu_sequence}
		\begin{tabular}{c|c|c}
			\hline\noalign{\smallskip}
			Method & Experimental Settings & Average Accuracy\\
			\hline
			HOG 3D~\cite{klaser2008spatio} & sequence-based & 70.63\%\\
			AdaLBP ~\cite{zhao2011facial}& sequence-based & 73.54\%\\
			Atlases~\cite{guo2012dynamic} & sequence-based & 75.52\%\\
			STM-ExpLet~\cite{liu2014learning}& sequence-based & 74.59\%\\
			DTAGN(Joint)~\cite{jung2015joint} & sequence-based & 81.46\%\\
			GoogLeNet (baseline) & image-based & 79.21\%\\
			PPDN (standard SGD) & image-based & 82.91\%\\
			\hline
			PPDN w/o peak  & image-based & \textbf{83.67\%}\\
			PPDN  & image-based & \textbf{84.59\%}\\
			\hline
		\end{tabular}
	\end{center}
\end{table}
\begin{table}[!tp]
 \setlength{\tabcolsep}{0.01pt}
 \setlength{\arraycolsep}{0.01pt} 
	\begin{center}
		\caption{Performance comparisons by adding the peak-piloted feature transformation on different convolutional layers when evaluated on Oulu-CASIA dataset.}
		\label{table:compare}
		\begin{tabular}{c|c|c|c|ccc}
			\hline\noalign{\smallskip}
			Method & inception layers & the last FC layer & the first FC layer & both FC layers\\
			\hline
			\footnotesize{Inception-3a} & \Checkmark & \XSolid  & \XSolid &\XSolid\\
			Inception-3b& \Checkmark  &\XSolid  & \XSolid &\XSolid\\
			Inception-4a& \Checkmark&\XSolid  & \XSolid  &\XSolid\\
			Inception-4b&\Checkmark &\XSolid  & \XSolid &\XSolid\\
			Inception-4c& \Checkmark &\XSolid  & \XSolid &\XSolid\\
			Inception-4d&\Checkmark &\XSolid  & \XSolid &\XSolid\\
			Inception-4e& \Checkmark &\XSolid  & \XSolid &\XSolid\\
			Inception-5a&\Checkmark &\XSolid  & \XSolid &\XSolid\\
			Inception-5b&\Checkmark & \XSolid & \XSolid &\XSolid\\
			Fc1&\Checkmark  & \XSolid&\Checkmark &\Checkmark\\
			Fc2& \Checkmark  & \Checkmark &\XSolid &\Checkmark  \\
			\hline
			Average Accuracy& 74.49\% & 73.33\%& 73.48\%& 74.99\%\\
			\hline
		\end{tabular}
	\end{center}

\end{table}
\setlength{\tabcolsep}{1.4pt}

\begin{table}[tp]
	\begin{center}
		\caption{Comparisons of the version with and without using peak information on Oulu-CASIA database in terms of average classification accuracy of the 10-fold cross-validation.}
		\label{table:auto}
		\begin{tabular}{c|c|c|c}
			\hline\noalign{\smallskip}
			Method & weak expression & peak expression & combined\\
			
			\hline
			PPDN w/o peak  & {67.52\%}& {83.79\%} & {74.01\%}\\
			PPDN  & \textbf{67.95\%}&\textbf{84.59\%} & \textbf{74.99\%}\\
			\hline
		\end{tabular}
	\end{center}

\end{table}

\setlength{\tabcolsep}{2pt}
\begin{table}[!tp]
	\begin{center}
		\caption{Face recognition rates for various poses under ``setting 1". }
		\label{table:face recognition}
		\begin{tabular}{c|c|c|c|c|c|c|c}
			\hline\noalign{\smallskip}
			Method  & $-45^{\circ}$& $-30^{\circ}$& $-15^{\circ}$& $+15^{\circ}$& $+30^{\circ}$& $+45^{\circ}$&Average\\

			\hline

			GoogLeNet (baseline)& 86.57\% & 99.3\% & 100\% & 100\%& 100\% &90.06\% & 95.99\%\\
			PPDN & \textbf{93.96\%}& \textbf{100\%} & \textbf{100\%} & \textbf{100\%} & \textbf{100\%} & \textbf{93.96\%} & \textbf{97.98\%}\\
			\hline
		\end{tabular}
	\end{center}
\end{table}

\setlength{\tabcolsep}{2pt}
\begin{table}[!tp]
	\begin{center}
		\caption{Face recognition rates for various poses under ``setting 2". }
		\label{table:face recognition2}
		\begin{tabular}{c|c|c|c|c|c|c|c}
			\hline\noalign{\smallskip}
			Method  & $-45^{\circ}$& $-30^{\circ}$& $-15^{\circ}$& $+15^{\circ}$& $+30^{\circ}$& $+45^{\circ}$&Average\\

			\hline

			Li et al.~\cite{li2012coupled}& 56.62\% & 77.22\% & 89.12\% & 88.81\%& 79.12\% &58.14\% & 74.84\%\\
			Zhu et al.~\cite{zhu2013deep}& 67.10\% & 74.60\% & 86.10\% & 83.30\%& 75.30\% &61.80\% & 74.70\%\\
			CPI~\cite{yim2015rotating}& 66.60\% & 78.00\% & 87.30\% & 85.50\%& 75.80\% &62.30\% & 75.90\%\\
			CPF~\cite{yim2015rotating}& 73.00\% & 81.70\% & 89.40\% & 89.50\%& 80.50\% &70.30\% & 80.70\%\\
			GoogLeNet (baseline)& 56.62\% & 77.22\% & 89.12\% & 88.81\%& 79.12\% &58.14\% & 74.84\%\\
			\hline
			PPDN & \textbf{72.06\%} & \textbf{85.41\%} & \textbf{92.44\%} & \textbf{91.38\%}& \textbf{87.07\%} &\textbf{70.97\%} & \textbf{83.22\%}\\
			\hline
		\end{tabular}
	\end{center}
\end{table}
\subsubsection{Comparisons with Sequence-based Approaches.} Unlike the 
still-image recognition setting, which evaluates the predictions of frames 
from a sequence, the sequence-based setting requires a prediction for the
whole sequence. Previous sequence-based approaches take the whole sequence 
as input and use motion information during inference. Instead, the PPDN 
regards each pair of non-peak and peak frame as input, and only outputs
the label of the peak frame as prediction for the whole sequence, in the 
testing phase. 
Tables~\ref{table:ck_sequence} and~\ref{table:oulu_sequence} compare
the PPDN to several sequence-based approaches plus ``GoogLeNet(baseline)'' 
on CK+ and Oulu-CASIA. Compared with~\cite{liu2014learning,liu2014deeply,jung2015joint}, which leverage motion information, the PPDN, which only relies on  
appearance information, achieves significantly better prediction performance.
On CK+, it has gains of $5.1\%$ and $2\%$ over 
`STM-ExpLet''~\cite{liu2014learning} and 
``DTAGN(Joint)''~\cite{jung2015joint}. On Oulu-CASIA it achieves
$84.59\%$ vs. the $75.52\%$ of ``Atlases''~\cite{guo2012dynamic} and 
the $81.46\%$ of ``DTAGN(Joint)''~\cite{jung2015joint}. In addition, we evaluate this experiment without peak information, i.e. selecting image with highest classification scores for all categories as peak frame in testing. PPDN achieves $99.2\%$ on CK+ and $83.67\%$ on Oulu-CASIA.

\subsubsection{PGS vs. standard SGD.} As discussed above, PGS suppresses gradients
from peak expressions, so as to drive the features of non-peak 
expression samples towards those of peak expression samples, but not the
contrary. Standard SGD uses all gradients, due to both non-peak and peak 
expression samples. We hypothesized that this will degrade recognition 
for samples of peak expressions, due to interference from non-peak 
expression samples. This hypothesis is confirmed by the results
of Tables~\ref{table:ck_compare} and~\ref{table:oulu_compare}. 
PGS outperforms standard SGD on all three test sets.
%

\subsubsection{Ablative Studies on Peak-Piloted Feature Transformation. }
The peak-piloted feature transformation, which is the key innovation of
the PPDN,  can be used on all layers of the network. Employing the 
transformation on different convolutional and fully-connected layers can 
result in different levels of supervision of non-peak responses by 
peak responses. For example, early convolutional layers extract fine-grained 
details (e.g., local boundaries or illuminations) of faces,
while later layers capture more semantic information, e.g., the appearance 
pattens of mouths and eyes. Table~\ref{table:compare} presents
an extensive comparison, by adding peak-piloted feature supervision on 
various layers. Note that we employ GoogLeNet~\cite{szegedy2015going}, 
which includes 9 inception layers, as basic network. Four different settings 
are tested: ``inception layers'' indicates that the loss of the peak-piloted 
feature transformation is appended for all inception layers plus the two 
fully-connected layers; ``the first FC layer,''``the last FC layer'' and 
``both FC layers'' append the loss to the first, last, and and both 
fully-connected layers, respectively. 

It can be seen that using the peak-piloted feature transformation only on the
two fully connected layers achieves the best performance. Using additional 
losses on all inception layers has roughly the same performance. Eliminating 
the loss of a fully-connected layer decreases performance by more
than 1\%. These results show that the peak-piloted feature transformation 
is more useful for supervising the highly semantic feature 
representations (two fully-connected layers) than the early convolutional 
layers.

\subsubsection{Absence of Peak Information.} Table~\ref{table:auto} 
demonstrates that the PPDN can also be used when the peak frame is not known
a priori, which is usually the case for real-world videos. Given all video 
sequences, we trained the basic ``GoogLeNet (baseline)'' with 10-fold 
cross validation. The models were trained with 9-folds and then
used to predict the ground-truth expression label in the remaining fold.
The frame with the highest prediction score in each sequence was treated as 
the peak expression image. The PPDN was finally trained using the
strategy of the previous experiments. This training procedure
is more applicable to videos where the information of the peak 
expression is not available. The PPDN can still obtain 
results comparable to those of the model trained with the ground-truth peak 
frame information.

\subsection{Generalization Ability of the PPDN}

The learning of the evolution from a hard sample to an 
easy sample is applicable to other face-related recognition 
tasks. We demonstrate this by evaluating the PPDN on face recognition. One 
challenge to this task is learning robust features,
invariant to pose and view. In this case,
near-frontal faces can be treated easy examples, similar to peak expressions 
in FER, while profile faces can be viewed as hard samples, similar to 
non-peak expressions. The effectiveness of PPDN in learning
pose-invariant features is demonstrated by comparing PPDN features
to the ``GoogLeNet(baseline)'' features on the popular Multi-PIE 
dataset~\cite{gross2010multi}. 

All the following experiments were conducted on the images of ``session 1''
on Multi-PIE, where the face images of 249 subjects are provided. Two 
experimental settings were evaluated to demonstrate the generalization 
ability of PPDN on face recognition. For the ``setting 1'' of 
Table~\ref{table:face recognition}, only images under normal illumination 
were used for training and testing, where  seven poses of the first 
100 subjects (ID from 001 to 100) were used for training and the
six poses (from $-45^{\circ}$ to  $45^{\circ}$) of the remaining individuals
used for testing. One frontal face per subject was used as gallery image. 
Overall, 700 images were used for training and 894 images for testing. 
By treating the frontal face and one of the profile faces as input, the 
PPDN can embed the implicit transformation from profile faces to frontal faces 
into the network learning, for face recognition purposes. In the
``setting 2'' of Table~\ref{table:face recognition2}, 100 subjects 
(ID 001 to 100) with seven different poses under 20 different illumination 
conditions were used for training and the rest with six poses and 
19 illumination conditions were used for testing. This led to 14,000 training
images and 16,986 testing images. Similarly to the first setting, PPDN takes 
the pair of a frontal face with normal illumination and one of the profile 
faces with 20 illuminations from the same subject as the input. 
The PPDN can thus learn the evolution from both the profile to the frontal 
face and non-normal to normal illumination. In addition to 
``GoogLeNet (baseline),'' we compared the PPDN to four state-of-the-art 
methods: controlled pose feature(CPF)~\cite{yim2015rotating}, controlled pose 
image(CPI)~\cite{yim2015rotating}, Zhu et al. ~\cite{zhu2013deep} and Li et 
al.~\cite{li2012coupled}. The pre-trained model, prepocessing steps, and 
learning rate used in the FER experiments were adopted here. Under 
``setting 1'' the network was trained with 10,000 iterations and under 
``setting 2'' with 30,000 iterations. Face recognition performance is 
measured by the accuracy of the predicted subject identity.

It can be seen that the PPDN achieves considerable improvements over 
``GoogLeNet (baseline)'' for the testing images of hard poses 
(i.e., $-45^{\circ}$ and $45^{\circ}$) in both ``setting 1'' and ``setting 2''. 
Significant improvements over ``GoogLeNet (baseline)'' are also observed
for the average over all poses ($97.98\%$ vs $95.99\%$ under ``setting 1'' 
and $83.22\%$ vs $74.84\%$ under ``setting 2''). 
The PPDN also beats all baselines by $2.52\%$ under ``setting 2''. 
This supports the conclusion that the PPDN can be effectively generalized to 
face recognition tasks, which benefit from embedding the evolution from 
hard to easy samples into the network parameters. 

\section{Conclusions}

In this paper, we propose a novel peak-piloted deep network for facial 
expression recognition. The main novelty is the embedding of the
expression evolution from non-peak to peak into the network parameters.
PPDN jointly optimizes an L2-norm loss of peak-piloted feature 
transformation and the cross-entropy losses of expression recognition. 
By using a special-purpose back-propagation procedure (PGS) for network 
optimization, the PPDN can  drive the intermediate-layer features of the 
non-peak expression sample towards those of the peak expression sample, 
while avoiding the inverse.\\  

\renewcommand{\theequation}{A-\arabic{equation}}
  \setcounter{equation}{0}  
  \section*{Appendix}  
  The loss 
  \begin{equation}
    J_1 = \sum_{i=1}^N \sum_{j \in \Omega}\Vert f_j(x^p_{i},W) - f_j(x^n_{i},W) \Vert^2 
 \end{equation}
 has gradient 
 \begin{equation}
 \begin{split}
   \nabla_W J_{1}& = 2\sum_{i=1}^N \sum_{j \in \Omega}(f_j(x^p_{i},W) - f_j(x^n_{i},W))\nabla_W f_j(x^n_i,W) \\
   &+ 2\sum_{i=1}^N \sum_{j \in \Omega}(f_j(x^p_{i},W) - f_j(x^n_{i},W))\nabla_W f_j(x^p_i,W).
 \end{split}
\end{equation}
 The PGS is 
 \begin{equation}
 \begin{split}
\widetilde{\nabla_W J_{1}}& = 2\sum_{i=1}^N \sum_{j \in \Omega}(f_j(x^p_{i},W) - f_j(x^n_{i},W))\nabla_W f_j(x^n_i,W)
\end{split}
  \end{equation}
  Defining
  \begin{equation}
    \begin{split}
      A = \sum_{i=1}^N \sum_{j \in \Omega}(f_j(x^p_{i},W) - f_j(x^n_{i},W))\nabla_W f_j(x^n_i,W)
    \end{split}
  \end{equation}\\
  and
  \begin{equation}
    \begin{split}
      B = \sum_{i=1}^N \sum_{j \in \Omega}(f_j(x^p_{i},W) - f_j(x^n_{i},W))\nabla_W f_j(x^p_i,W)
    \end{split}
  \end{equation}
  it follows that
  \begin{equation}
 \begin{split}
   <\nabla_W J_{1},\widetilde{\nabla_W J_{1}}>& = -4 <A,B> + 4\Vert A \Vert^2 
 \end{split}
  \end{equation}
 or \begin{equation}
 \begin{split}
<\nabla_W J_{1},\widetilde{\nabla_W J_{1}}>& = -4 \Vert A\Vert \Vert B\Vert \cos{\theta} + 4\Vert A \Vert^2 
\end{split}
  \end{equation}
  where $\theta$ is angle between A and B. 
  Hence, the dot-product is greater than zero when
  \begin{equation}
    \begin{split}
      \Vert B\Vert \cos{\theta} < \Vert A \Vert.
    \end{split}
  \end{equation}
  This holds for sure as $\nabla_W f_j(x^n_i,W)$ converges to 
  $\nabla_W f_j(x^p_i,W)$ which is the goal of optimization, but is 
  generally true if the sizes of gradients $\nabla_W f_j(x^n_i,W)$ and 
  $\nabla_W f_j(x^p_i,W)$ are similar on average. Since the dot-product is positive, $\widetilde{\nabla_W J_{1}}$  is a descent (although not a steepest descent) direction for the loss function $J_1$. Hence, the PGS is a descent direction for the total loss. Note that, because there are also the gradients of $J_2$ and $J_3$, this can hold even when (A-8) is violated, if the gradients of J2 and J3 are dominant. Hence, the PGS is likely to converge to a minimum of the loss. 
\clearpage
\bibliographystyle{splncs}
\bibliography{egbib}
\clearpage

\end{document}